%% file: iclr2027_conference.tex
\documentclass{article} 
\usepackage{iclr2027_conference,times}

\input{math_commands.tex}

\usepackage{hyperref}
\usepackage{url}
\usepackage{booktabs}
\usepackage{multirow}
\usepackage{graphicx}
\usepackage{amsmath,amssymb}
\usepackage{enumitem}
\usepackage{xcolor}
\usepackage{tikz}
\usetikzlibrary{arrows.meta,positioning,calc,fit,backgrounds,shapes.misc}
\usepackage{algorithm}
\usepackage{algpseudocode}

\usepackage{titlesec}
\titlespacing*{\paragraph}{0pt}{0pt plus 0.2pt minus 0.2pt}{0.5em}

\usepackage{xspace}
\makeatletter \DeclareRobustCommand\onedot{\futurelet\@let@token\@onedot}
\def\@onedot{\ifx\@let@token.\else.\null\fi\xspace} \def\eg{\emph{e.g}\onedot}
 \def\ie{\emph{i.e}\onedot}

\makeatother

\definecolor{ourscol}{RGB}{192,57,43}
\definecolor{trmcol}{RGB}{110,110,110}

\newcommand{\ours}{TRM4CO}

\title{Compute Time Scaling with Recursive Models for Combinatorial
Optimization}

\author{Zhengxi Zhang \& Paul Swoboda \\
Heinrich Heine University D\"usseldorf \\
\texttt{\{zhengxi.zhang,paul.swoboda\}@hhu.de}}

\iclrfinalcopy 
\begin{document}

\maketitle
\lhead{}  

\begin{abstract}
We propose Tiny Recursive Models for Combinatorial Optimization (\ours{}), a
general neural method for combinatorial optimization that scales both depth (how
often we recursively invoke our network) and width (how much we sample in
parallel).
Both are fundamental for combinatorial optimization: hard instances demand a
large amount of compute, while a small network is essential to avoid overfitting
and capture the algorithmic essence of optimization.
In particular, our method consists of a graph-aware tiny recursive model that
iterates on a latent state with adaptive halting and needs only a lightweight
problem-specific decoder.
Compared with previous heatmap-based general neural solvers, it achieves a
better balance between solution quality and inference speed on both the
Traveling Salesman Problem~(TSP) and the Maximum Independent Set~(MIS) problem,
and remains competitive with hybrid methods that combine neural components with
heuristics specific to each problem.
With the same backbone architecture for both tasks, \ours{} outperforms every
diffusion-based solver on TSP from 500 to 10,000 cities at a lower inference
cost, and on the standard Erd\H{o}s--R\'enyi-[700-800] MIS benchmark it
surpasses all neural solvers except those that only work well on MIS.
We then explore self-relabeling for self-supervised training.
We periodically replace the current set of training labels with the model's own
better solutions, as an alternative training signal.
Self-relabeling can, while forgoing supervision from near-optimal solutions,
still result in on-par quality.

\end{abstract}

\section{Introduction}
\label{sec:intro}

Deep learning for combinatorial optimization faces a mismatch.
Most neural networks have large numbers of parameters, yet do relatively little
computation per parameter.
Combinatorial optimization algorithms, on the other hand, are comparatively
short programs that perform a lot of computation.
It is an open problem to design architectures with large and deep computational
graphs, which give the representational capacity that optimization demands, from
relatively few parameters, which keep them fast and prevent overfitting.
One answer to this is compute time scaling.
To this end, we bring current advances in recursive deep learning architectures
to neural combinatorial optimization.

Specifically, we propose \ours{}, Tiny Recursive Model for Combinatorial
Optimization.
Our backbone is the Tiny Recursive Model~(TRM, \citealp{jolicoeur2025trm}), a
two-block Transformer with a few million parameters that is applied repeatedly
to a latent state and an evolving answer.
It is trained with deep supervision to improve the answer at every step.
\ours{} scales the test-time compute of the TRM recursion along two axes
(Figure~\ref{fig:overview}): along depth, \ie the number of times the network is
invoked, and along width, \ie the number of solutions it samples in parallel.
The network trains either on solver labels or without any, in a self-supervised
loop that samples solutions, keeps those that improve on the stored ones, trains
on them, and repeats this hill-climbing.

Existing work on neural combinatorial optimization scales compute differently.
Autoregressive constructive policies
\citep{vinyals2015pointer,kool2019attention,kwon2020pomo,luo2023lehd,drakulic2023bqnco}
grow a solution one node at a time, which is accurate but sequential, so at
thousands of nodes they rely in practice on decomposition
\citep{ye2024glop,zheng2024udc,li2025drhg}.
Heatmap-based solvers instead predict all decision variables at once and hand
the rest to a search.
Among them diffusion solvers
\citep{sun2023difusco,li2023t2t,li2024fastt2t,zhao2025disco,feng2026cam} also
iterate, but along a fixed noise schedule.
The variants that scale test-time compute do so in heatmap space
\citep{li2023t2t,li2024fastt2t}, where a search is still required to obtain a
solution.
Other heatmap solvers spend the extra compute around the network rather than in
it, by fine-tuning a meta-learned policy on each instance and sampling massively
in parallel \citep{qiu2022dimes}, or by learning an optimizer whose iterations
update the heatmap \citep{dernedde2025moco}.
Earlier approaches train a graph neural network~(GNN) as a supervised edge
classifier \citep{joshi2019gcn,fu2021attgcn} and are no longer competitive on
their own.

We benchmark on the Maximum Independent Set~(MIS) problem and the Traveling
Salesman Problem~(TSP), two prototypical combinatorial optimization problems
in the literature.
To make recursive models work on both, we augment them with (i)~an embedding of
the problem structure (adjacency matrices modulating attention via a learnable
per-head bias for MIS, embedded coordinates for TSP), (ii)~lightweight
problem-specific decoders (peeling for MIS, edge insertion with 2-opt for TSP),
and (iii)~halting and selection driven by the decoded solutions. 
The network size stays constant by design, so additional compute goes into
iterations rather than parameters.

Empirically, \ours{} reaches the lowest optimality gap of all learned solvers on
TSP-500 and TSP-1000, \eg 0.11\% on TSP-500 in 2.1 seconds per instance, and
matches the best diffusion solver on TSP-10000.
On MIS, solution quality grows with test-time compute, and with 200 rollouts
\ours{} surpasses every general-purpose neural solver and the KaMIS reference on
Erd\H{o}s--R\'enyi~(ER) graphs with 700--800 nodes.
Finally, self-relabeling raises 16\% of the training labels above those of KaMIS
and yields our best MIS checkpoint.
Without any solver labels, the same loop trains an MIS model from scratch whose
own labels match or beat KaMIS on 42\% of the training graphs and whose test gap
comes close to that of our best checkpoint (Section~\ref{sec:selfsup}).

In summary, our contributions are as follows:
\begin{itemize}[topsep=0pt, itemsep=0pt, parsep=0pt, partopsep=0pt]
    \item We propose a deep recursive architecture for combinatorial
      optimization that can efficiently scale along depth and width.
    \item Our method can be trained in a supervised way from solver labels or in
      a self-supervised manner by self-relabeling, \ie generating multiple
      labels and hill-climbing on the best ones.
    \item We show experimentally on MIS and TSP that we match or outperform
      state-of-the-art neural combinatorial optimization approaches.
\end{itemize}

\section{Method}
\label{sec:method}

\begin{figure}[t]
\begin{center}
\resizebox{\textwidth}{!}{\input{figures/framework}}
\end{center}
\caption{\ours{} overview.
An encoder turns an instance into node tokens plus $P$ prefix tokens and, for
MIS, the adjacency matrix $A$ that biases every attention call of the shared
recursive backbone (Eq.~\ref{eq:edgebias}).
Each step runs one TRM recursion, and test-time compute scales along depth $D$,
the number of steps, and width $K$, the number of parallel rollouts.
A problem-specific decoder turns the answer state of every step into a feasible
solution, with sequential peeling for MIS and edge insertion with 2-opt for TSP,
and the objective keeps the best candidate.
In training, a solved check against the label halts an instance, and better
solutions can replace the labels.
Red marks what we add to TRM.}
\label{fig:overview}
\end{figure}
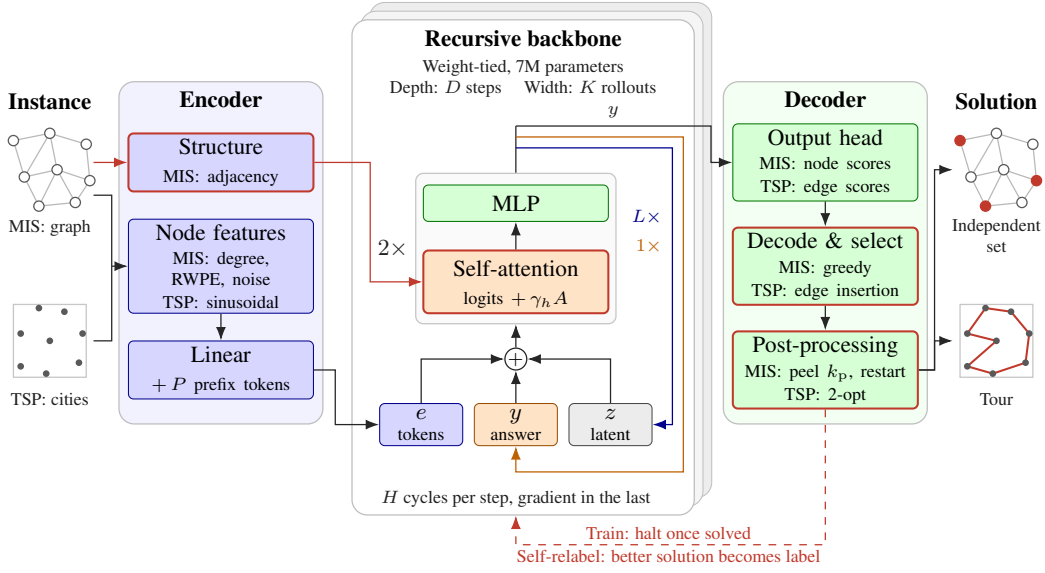

At a high level, our method works as follows (Figure~\ref{fig:overview}).
The \underline{encoder} turns a problem instance, which comes with a set of
feasible solutions and an objective, into tokens.
A problem-agnostic \underline{recursive backbone} then updates a latent state
and an answer state iteratively.
We scale the compute of the backbone along depth, the number of steps, and along
width, the number of rollouts.
We also inject noise into the latents for diversity.
Periodically, a \underline{decoder} reads the answer state and turns it into a
solution.
At inference, we keep the best solution over all readings across depths and
rollouts.
Optionally, during training, we replace the label of an instance whenever we
sample a better solution.

We explain all components in detail below.

\subsection{Background: Tiny Recursive Model (TRM)}
\label{sec:trm}

We use the Tiny Recursive Model \citep{jolicoeur2025trm} as the backbone.
It is a simplification of the Hierarchical Reasoning Model
(HRM, \citealp{wang2025hrm}).
TRM keeps two states of the same shape as the input tokens $e$.
The latent $z$ carries the reasoning and the answer $y$ carries the current
prediction.
As shown in Figure~\ref{fig:recursion} (Appendix~\ref{app:alg}), one weight-tied
network $f_\theta$ of two Transformer blocks updates both states in cycles:
\begin{equation}
  \begin{aligned}
    z &\leftarrow f_\theta(z + y + e) \qquad && (L \text{ times}), \\
    y &\leftarrow f_\theta(z + y) && (\text{once}),
  \end{aligned}
  \label{eq:trm}
\end{equation}
and one recursion step runs $H$ such cycles, of which only the last carries
gradient.
Deep supervision chains up to $D$ recursion steps, carrying the states $(y, z)$
detached from one step to the next and applying the loss to the output head
$g_\theta(y)$ after each step.
Training memory is thus constrained to one cycle, while the effective depth is
$2H(L+1)D$ blocks.
A learned halting head on $y$ decides during training when an instance stops
recursing \citep{graves2016act}.
This acts as hard mining, because easy instances leave early and the remaining
steps go to hard ones.
Adding noise to $z$ at test time and running several copies in parallel yields
extra samples \citep{sghaier2026ptrm}.

\subsection{The \ours{} framework}
\label{sec:framework}

\paragraph{Input embedding and encoder.}
The encoder maps an instance $x$ to $n$ node tokens in $\mathbb{R}^{d}$ and, for
MIS, passes on the adjacency matrix $A \in \{0,1\}^{n \times n}$
(Table~\ref{tab:hooks}).
$P$ learnable prefix tokens are prepended and serve as global registers, \ie
tokens attached to no node that every node can attend to, and we write $e$ for
the resulting $P + n$ tokens.
$A$ is zero-padded on the prefix rows and columns, so those registers receive no
bias.
No global positional encoding over the token order is used for either problem,
so the model is permutation equivariant by construction.
For MIS, the node tokens are a linear projection of the degree, a random-walk
positional encoding~(RWPE, \citealp{dwivedi2022lspe}), and a uniform random
scalar that breaks the symmetry between structurally identical nodes.
The edges enter the recursion through $A$.
For TSP, the geometry is entirely in the coordinates $X$.
The TSP node tokens are a linear projection of $X$ lifted with sinusoidal
features at $L_{\mathrm{pe}}$ octaves per axis \citep{mildenhall2020nerf}, and
no attention bias is needed.
This asymmetry is deliberate.
The graph structure of MIS has to be injected into attention, whereas for TSP on
the unit square, the coordinates already determine locality and the model can
recover it via standard self-attention.

\paragraph{Recursive core with edge-biased attention.}
The core of our model is the two-block network of Eq.~(\ref{eq:trm}) with one
change.
Every attention head $h$ of every call computes
\begin{equation}
  \mathrm{softmax}\!\left( \frac{\mathbf{Q}_h \mathbf{K}_h^\top}{\sqrt{d_h}} + \gamma_h A \right),
  \label{eq:edgebias}
\end{equation}
where $\mathbf{Q}_h$ and $\mathbf{K}_h$ are the queries and keys of head $h$
with dimension $d_h$, and $\gamma_h$ is a learnable scalar per head and block,
initialized at zero.
$A$ is the adjacency matrix of the instance, with $A_{ij} = 1$ when nodes $i$
and $j$ are neighbors and $0$ otherwise, so the bias raises or lowers the
attention between adjacent nodes.
With $\gamma_h = 0$ the head is standard self-attention, as
$\gamma_h \to \infty$ it attends only to neighbors, and with $\gamma_h < 0$ it
is biased toward non-neighbors.
Each head therefore learns its own mixture of local and global attention
\citep{ying2021graphormer} at the cost of one scalar per head.
For MIS, this stands in for a separate message-passing GNN encoder
(Table~\ref{tab:ablation}).
The recursive core has about 7M parameters, and its size does not change between
problems.

\paragraph{Depth and width.}
Depth $D$ is the number of recursion steps.
Since the output head $g_\theta$ is applied after every step, one run of depth
$D$ yields $D$ candidate solutions.
Width $K$ perturbs the latent with Gaussian noise $z \leftarrow z + \sigma \xi$,
$\xi \sim \mathcal{N}(0, I)$, at every step \citep{sghaier2026ptrm} and runs $K$
copies of the instance in one batched forward.
Every candidate is feasible by construction, and its objective $c$ is evaluated,
so the best of them is selected by exact ranking.
The same selection procedure can also produce training labels
(Section~\ref{sec:algorithm}).
Solution quality improves along both axes, and the cost is linear in
$K \times D$.
Both axes therefore scale compute time on the same weights, with no additional
parameters or training.

\paragraph{Decoding and selection.}
An output head $g_\theta$ reads the answer state.
For MIS, the head is a per-node two-way classifier, and the difference between a
node's positive and negative logits is its score.
For TSP, the head computes a bilinear successor score
$g_{ij} = \langle W_q y_i, W_k y_j \rangle / \sqrt{d_s}$ over ordered node
pairs, with $d_s$ the dimension of the projected vectors.
A lightweight decoder turns the scores into a feasible solution, and the
objective $c$ evaluates it for selection.
For MIS, the per-step decoder is greedy and also serves as the solved check in
training (Section~\ref{sec:halting}).
Nodes are scanned in order of decreasing score, and a node is added whenever
none of its neighbors has been picked.
Optionally, Gumbel noise of temperature $\tau$ perturbs the scores.
We wrap the recursion in \emph{sequential peeling}.
After $D$ steps, up to $k_{\mathrm{p}}$ nodes are committed, taken in order of
decreasing score with conflicting nodes skipped.
The closed neighborhoods of the committed nodes are masked, and the recursion
restarts on the residual graph until it is empty.
Committing $k_{\mathrm{p}}$ nodes each time costs about $|S|/k_{\mathrm{p}}$
rounds of $D$ steps, with $|S|$ the final set size, and improves the solution
substantially over one-shot decoding (Table~\ref{tab:ablation}).
We decode TSP by greedy edge insertion as in \citet{sun2023difusco}.
Candidate edges are ranked by $\bar g_{ij} / d_{ij}$, where $\bar g_{ij}$ is the
symmetrized successor probability from a row-wise softmax of $g$ and $d_{ij}$
the Euclidean distance.
They are inserted in rank order while respecting the degree and subtour
constraints, until a tour is formed.
The tour selected at the end is repaired once with 2-opt.

\subsection{Training and inference}
\label{sec:algorithm}

Training is deep supervision over a batch of instances.
Each slot of the batch carries one instance across steps and is refilled once
that instance halts.
Each step applies a cross-entropy loss between $g_\theta(y)$ and the label, per
node for MIS and per successor for TSP.
Inference runs $D$ recursion steps on $K$ noisy copies, decodes and scores after
every step, and returns the best-scoring solution.
A problem-specific wrapper completes the result.
For TSP, it repairs the obtained tour with 2-opt.
For MIS, it wraps the recursion in sequential peeling, which replaces the
per-step selection.
Each of the $K$ copies peels independently, and the objective compares the $K$
finished sets at the end.
Algorithm~\ref{alg:trm4co} in Appendix~\ref{app:alg} collects these procedures,
which MIS and TSP share, with the problem-specific hooks of
Table~\ref{tab:hooks} plugged in.

\paragraph{Halting on a solved check.}
\label{sec:halting}
TRM trains its halting head to predict whether the current answer matches the
label token by token, and halts on the head's decision.
We drop this signal in favor of the problem-specific check in
Table~\ref{tab:hooks} and halt on it directly, without a learned head.
For MIS, the greedy decoding of the current scores must reach at least the label
size, which credits any of the many optima a graph may have.
For TSP, the label tour is oriented counterclockwise so that every city has a
unique successor.
The check requires the predicted successors to agree with the label exactly.
An instance halts when the check passes or when it exhausts its budget of $D$
steps.
The depth an instance receives is therefore adaptive during training.
At test time, the depth is set explicitly by $D$, and selection by the objective
$c$ takes the place of the halting decision.
Since every step applies the same set of weights, evaluation may run deeper than
training.

\paragraph{Self-relabeling.}
\label{sec:selfrelabel}
Optionally, we let the model relabel its own training set.
Every $T$ epochs, the model solves every training instance, and the solution
replaces the current label whenever it is strictly better, \ie a larger
independent set or a shorter tour.
The labels therefore only ever improve.
The same loop can also train from random initialization without any external
solver.
One pass of self-relabeling with the untrained model seeds every label before
training starts, and every later pass tightens the labels.
Section~\ref{sec:selfsup} reports the results for the self-supervised settings.

\section{Experiments}
\label{sec:exp}

\subsection{Setup}
\label{sec:setup}

\paragraph{Benchmarks.}
We follow the training and test data splits of DIFUSCO \citep{sun2023difusco}.
For MIS, these are Erd\H{o}s--R\'enyi~(ER) graphs \citep{erdos1960evolution}
with 700 to 800 nodes and edge probability 0.15 (ER-[700-800], 16,384 training
and 128 test graphs), and graphs obtained by reducing the CNF formulas of SATLIB
\citep{hoos2000satlib} to MIS, with up to 1347 nodes (39,500 training and 500
test graphs).
Both benchmarks come from \citet{qiu2022dimes}.
For TSP, we generate 128,000, 64,000 and 6,400 training instances with 500, 1000
and 10000 cities respectively, uniformly distributed in the unit square, and we
adopt the TSP-500/1000/10000 test sets of \citet{fu2021attgcn} with 128, 128 and
16 instances.
We compute all training labels ourselves, with KaMIS
\citep{lamm2017kamis,hespe2019kernelization} at a 10-second budget per graph for
ER, Gurobi \citep{gurobi2026} for SATLIB, and Concorde
\citep{applegate2006concorde} or LKH-3 \citep{helsgaun2017lkh3} for TSP
(Appendix~\ref{app:data}).
The training setup is detailed in Appendix~\ref{app:hyper}.


\paragraph{Baselines.}
Tables~\ref{tab:mis} and~\ref{tab:tsp} divide the compared methods into three
groups: general-purpose neural solvers that report results on both MIS and TSP
(heatmap, diffusion and divide-and-conquer solvers), learning-free methods
(samplers and heatmap-free search), and neural solvers specific to MIS or TSP.
Classical solvers (KaMIS, Concorde and LKH-3) serve as references.

\paragraph{Metrics and runtime.}
For MIS, we report the mean independent set size and the optimality gap to
KaMIS.
For TSP, we report the mean tour length and the gap against Concorde at
$n = 500$ and $1000$ and against LKH-3 at $n = 10000$, as in DIFUSCO and its
successors.
Quality numbers are taken from the original papers, except for the reruns listed
in Appendix~\ref{app:tablenotes}.
We measure only the runtime, on one RTX 4090 with each method's released code
and checkpoints, from input to final solution (Appendix~\ref{app:protocol}).
The reference solvers are listed with the CPU times reported in the original
papers.

\subsection{Main results on MIS}
\label{sec:mis}

\begin{table}[!t]
\caption{MIS results.
We list mean independent set size on ER-[700-800] and SATLIB, gap in \% with
respect to the KaMIS reference, and seconds per instance on one RTX 4090.
Type: S = supervised, U = unsupervised, RL = reinforcement learning (details in
Appendix~\ref{app:tablenotes}). ``--'' means that the number is not available,
and ``$^\dagger$'' marks evaluation on the method's own instances from the same
distribution.
The KaMIS time is measured on CPU.
Bold marks the lowest gap on each benchmark.}
\label{tab:mis}
\centering
\footnotesize
\setlength{\tabcolsep}{5.8pt}
\begin{tabular}{llrrr@{\hspace{6pt}}rrr}
\toprule
\multicolumn{2}{l}{Dataset} & \multicolumn{3}{c}{ER-[700-800]} & \multicolumn{3}{c}{SATLIB} \\
\cmidrule(lr){3-5} \cmidrule(lr){6-8}
Method & Type & Size & Gap \% & s/inst & Size & Gap \% & s/inst \\
\midrule
KaMIS (reference) & heuristic & 44.87 & 0.00 & 24 & 425.96 & 0.00 & 4.5 \\ 
\midrule
\multicolumn{8}{l}{\emph{General-purpose neural solvers (MIS and TSP)}} \\
DIMES \citep{qiu2022dimes} & RL & 42.06 & 6.26 & -- & 423.28 & 0.63 & -- \\ 
DIFUSCO \citep{sun2023difusco} & S & 41.12 & 8.36 & 7.4 & 425.13 & 0.19 & 1.6 \\ 
T2T \citep{li2023t2t} & S + search & 41.37 & 7.80 & 9.0 & 425.22 & 0.17 & 1.5 \\ 
UDC \citep{zheng2024udc} & RL & 42.88 & 4.44 & 6.9 & -- & -- & -- \\ 
Fast T2T \citep{li2024fastt2t} & S + search & 41.73 & 7.00 & 0.92 & 425.23 & 0.17 & 1.4 \\ 
Moco \citep{dernedde2025moco} & U + search & 44.87 & 0.00 & 22.8 & 425.02 & 0.22 & 109 \\ 
DISCO \citep{zhao2025disco} & S & 42.21 & 5.93 & -- & 425.06 & 0.21 & -- \\ 
COExpander \citep{ma2025coexpander} & S & 42.56 & 5.15 & 1.2 & 425.28 & 0.16 & 1.3 \\ 
COExpander + RLSA & S + sampler & \textbf{45.27} & \textbf{-0.89} & 4.5 & 425.45 & 0.12 & 8.6 \\ 
CADO \citep{song2026cado} & S + RL & 43.62 & 2.79 & -- & 425.43 & 0.12 & -- \\ 
CAM \citep{feng2026cam} & U & 44.16 & 1.58 & 0.2 & -- & -- & -- \\ 
HyCO \citep{li2026hyco} & S + RL & 42.38\rlap{$^\dagger$} & 5.55\rlap{$^\dagger$} & -- & 425.40\rlap{$^\dagger$} & 0.13\rlap{$^\dagger$} & -- \\ 
\midrule
\multicolumn{8}{l}{\emph{Learning-free methods}} \\
iSCO \citep{sun2023isco} & sampler & 45.15 & -0.62 & 55 & 424.16 & 0.42 & 87 \\ 
ReSCO \citep{li2025resco} & sampler & 45.24 & -0.82 & 73 & 424.21 & 0.41 & 258 \\ 
RLSA \citep{feng2025rlnn} & sampler & 45.05 & -0.40 & 0.65 & -- & -- & -- \\ 
\midrule
\multicolumn{8}{l}{\emph{MIS-specific neural solvers (no TSP results)}} \\
LwD \citep{ahn2020lwd} & RL & 41.17 & 8.25 & -- & 424.80\rlap{$^\dagger$} & 0.27\rlap{$^\dagger$} & -- \\ 
GFlowNets \citep{zhang2023gflownets} & U & 41.14\rlap{$^\dagger$} & 8.31\rlap{$^\dagger$} & -- & 423.54\rlap{$^\dagger$} & 0.57\rlap{$^\dagger$} & -- \\ 
DiffUCO \citep{sanokowski2024diffuco} & U & 43.98 & 1.98 & -- & -- & -- & -- \\ 
SDDS \citep{sanokowski2025sdds} & U & 43.31 & 3.48 & -- & -- & -- & -- \\ 
RLNN \citep{feng2025rlnn} & U & 43.34 & 3.41 & 0.26 & -- & -- & -- \\ 
DDEA \citep{salvasoler2026ddea} & S + evolution & 44.13 & 1.65 & 396 & -- & -- & -- \\ 
\midrule
\multicolumn{8}{l}{\emph{\ours{} (general-purpose; noise $\sigma{=}0.2$; peel $k_{\mathrm{p}}{=}2$, Gumbel noise $\tau{=}0.5$ on ER; one-shot on SATLIB)}} \\
\ours{} ($K{=}10$, $D{=}32$) & S + self & 44.37 & 1.11 & 18.7 & \textbf{425.96} & \textbf{0.00} & 2.2 \\ 
\ours{} ($K{=}100$, $D{=}64$) & S + self & 44.93 & -0.13 & 368 & -- & -- & -- \\ 
\ours{} ($K{=}200$, $D{=}32$) & S + self & 45.04 & -0.38 & 373 & -- & -- & -- \\ 
\bottomrule
\end{tabular}
\end{table}

Table~\ref{tab:mis} compares quality and cost.
On ER-[700-800], the learned solvers span a wide range.
The supervised diffusion solvers have the highest gaps, the recent RL and
unsupervised solvers improve on them, and Moco matches the reference at the
price of hundreds of heatmap-search steps per graph.
\ours{} runs the same weights at every depth and width.
At a moderate width it sits among the strongest single-model solvers, and with
more rollouts it passes Moco and the KaMIS reference, as the only learned solver
without a sampler to do so.
The learning-free samplers reach slightly lower gaps still.
RLSA, alone and on top of COExpander, dominates on ER graphs, which confirms the
observation of \citet{sun2023isco} that these graphs reward long search more
than learning.
\ours{} closes most of that distance with a general recipe, but at a much higher
cost, since peeling reruns the recursion about $|S| / k_{\mathrm{p}}$ times.
SATLIB shows the recursion without peeling.
All neural solvers are within 1\% of KaMIS there, and \ours{} at $K = 10$ with
one-shot decoding matches the reference and solves almost every instance
exactly, in the cost band of the diffusion solvers with sampling.

\subsection{Main results on TSP}
\label{sec:tsp}

\begin{table}[!t]
\caption{TSP results on uniform instances.
We report mean tour length, gap in \% against Concorde ($n = 500, 1000$) and
LKH-3 ($n = 10000$), and seconds per instance on one RTX 4090.
The Search column indicates the search procedure that produces each result,
either post-processing after the network or a custom refinement loop (details in
Appendix~\ref{app:tablenotes}).
The meanings of ``--'', ``$^\dagger$'', and the reference solver time follow
those of Table~\ref{tab:mis}.
Bold marks the lowest gap at each problem size.}
\label{tab:tsp}
\centering
\footnotesize
\setlength{\tabcolsep}{1.35pt}
\begin{tabular}{llrrr@{\hspace{8pt}}rrr@{\hspace{8pt}}rrr}
\toprule
\multicolumn{2}{l}{Dataset} & \multicolumn{3}{c}{TSP-500} & \multicolumn{3}{c}{TSP-1000} & \multicolumn{3}{c}{TSP-10000} \\
\cmidrule(lr){3-5} \cmidrule(lr){6-8} \cmidrule(lr){9-11}
Method & Search & Len. & Gap \% & s/inst & Len. & Gap \% & s/inst & Len. & Gap \% & s/inst \\
\midrule
Concorde / LKH-3 (reference) & -- & 16.55 & 0.00 & 18 & 23.12 & 0.00 & 187 & 71.77 & 0.00 & 1980 \\ 
\midrule
\multicolumn{11}{l}{\emph{General-purpose neural solvers (MIS and TSP)}} \\
DIMES \citep{qiu2022dimes} & MCTS & 16.84 & 1.76 & 67 & 23.69 & 2.46 & 136 & 74.06 & 3.19 & 1219 \\ 
DIFUSCO \citep{sun2023difusco} & 2-opt & 16.65 & 0.57 & 2.9 & 23.45 & 1.43 & 12.4 & 73.89 & 2.95 & 266 \\ 
DIFUSCO & MCTS & 16.63 & 0.46 & 51 & 23.39 & 1.17 & 104 & 73.62 & 2.58 & 1076 \\ 
T2T \citep{li2023t2t} & 2-opt & 16.61 & 0.37 & 4.4 & 23.30 & 0.78 & 11.0 & 73.81 & 2.84 & -- \\ 
UDC \citep{zheng2024udc} & conquer & 16.82 & 1.66 & 1.8 & 23.53 & 1.78 & 2.7 & 74.71 & 4.09 & 1.9 \\ 
Fast T2T \citep{li2024fastt2t} & 2-opt & 16.58 & 0.21 & 2.5 & 23.22 & 0.42 & 5.9 & 72.90 & 1.57 & -- \\ 
Moco \citep{dernedde2025moco} & 2-opt & 16.62 & 0.40 & 98 & -- & -- & -- & -- & -- & -- \\ 
DISCO \citep{zhao2025disco} & 2-opt & -- & -- & -- & -- & -- & -- & 73.81 & 2.84 & -- \\ 
COExpander \citep{ma2025coexpander} & 2-opt & 16.59 & 0.25 & 1.0 & 23.27 & 0.64 & 2.2 & 72.80 & 1.45 & 25 \\ 
CADO \citep{song2026cado} & 2-opt & 16.65 & 0.61 & -- & 23.32 & 0.88 & -- & 73.69 & 2.68 & -- \\ 
CAM \citep{feng2026cam} & 2-opt & 16.91 & 2.18 & 3.2 & 23.70 & 2.51 & 12.6 & -- & -- & -- \\ 
HyCO \citep{li2026hyco} & 2-opt & 16.74 & 1.17 & -- & 23.64 & 2.26 & -- & -- & -- & -- \\ 
\midrule
\multicolumn{11}{l}{\emph{Learning-free methods}} \\
iSCO \citep{sun2023isco} & sampler & 16.64 & 0.54 & -- & 23.33 & 0.91 & -- & 74.02 & 3.14 & -- \\ 
SoftDist \citep{xia2024rethinking} & MCTS & 16.78 & 1.44 & 50 & 23.63 & 2.20 & 101 & 74.03 & 3.13 & 1012 \\ 
Zero heatmap \citep{pan2026beyondheatmap} & MCTS & 16.66 & 0.66 & 50 & 23.39 & 1.16 & 100 & 74.50 & 3.79 & 102 \\ 
GT-Prior \citep{pan2026beyondheatmap} & MCTS & 16.63 & 0.50 & 50 & 23.31 & 0.85 & 100 & 73.31 & 2.13 & 101 \\ 
\midrule
\multicolumn{11}{l}{\emph{TSP-specific neural solvers (no MIS results)}} \\
Att-GCN \citep{fu2021attgcn} & MCTS & 16.97 & 2.54 & 20 & 23.86 & 3.22 & 40 & 74.93 & 4.39 & 401 \\ 
BQ-NCO \citep{drakulic2023bqnco} & beam search & 16.64 & 0.55 & 3.5 & 23.44 & 1.38 & 8.6 & -- & -- & -- \\ 
UTSP \citep{min2023utsp} & local search & 16.68 & 0.84 & 20 & 23.39 & 1.18 & 41 & -- & -- & -- \\ 
LEHD \citep{luo2023lehd} & RRC & 16.57 & 0.16 & 53.5 & 23.29 & 0.75 & 256 & -- & -- & -- \\ 
GLOP \citep{ye2024glop} & revisers & 16.91 & 1.99 & 0.8 & 23.84 & 3.11 & 1.6 & 75.29 & 4.90 & 4.7 \\ 
SIL \citep{luo2024sil} & PRC & -- & -- & -- & 23.31 & 0.81 & 349 & 73.32 & 2.00 & 1011 \\ 
DualOpt \citep{zhou2025dualopt} & LKH-3 + NN & -- & -- & -- & 23.31 & 0.83 & 2.0 & 72.62 & 1.18 & 13.9 \\ 
DRHG \citep{li2025drhg} & LNS & 16.58 & 0.23 & 25.8 & 23.19 & 0.29 & 52 & 72.85 & 1.33 & 106 \\ 
HyperNS \citep{lu2026hyperns} & LK search & -- & -- & -- & 23.20 & 0.35 & -- & \textbf{72.44} & \textbf{0.93} & -- \\ 
PCI \citep{basson2026pci} & 2-opt & 16.57\rlap{$^\dagger$} & 0.17\rlap{$^\dagger$} & -- & 23.19\rlap{$^\dagger$} & 0.31\rlap{$^\dagger$} & -- & -- & -- & -- \\ 
\midrule
\multicolumn{11}{l}{\emph{\ours{} (general-purpose; noise $\sigma{=}0.2$)}} \\
\ours{} ($K{=}40$, $D{=}32$) & 2-opt & \textbf{16.56} & \textbf{0.11} & 2.1 & \textbf{23.18} & \textbf{0.27} & 4.4 & 72.81 & 1.45 & 364 \\ 
\bottomrule
\end{tabular}
\end{table}

Table~\ref{tab:tsp} compares gaps and cost at 500, 1000, and 10000 cities.
At $n = 500$ and $1000$, \ours{} has the lowest gap of all learned solvers, at a
cost within the band of diffusion solvers.
Search matters in these pipelines.
DIFUSCO with MCTS improves on its 2-opt variant only through the search, and the
heatmap-free controls of \citet{pan2026beyondheatmap} land in the same range
under MCTS, at more than $20\times$ our cost per instance.
Learning-free methods fare worse on TSP than on MIS, where samplers alone beat
the KaMIS reference (Table~\ref{tab:mis}).
An ER graph has many maximum independent sets, whereas a TSP instance on the
unit square almost surely has a unique optimal tour, so a search without learned
guidance has many targets on MIS, but only a single one on TSP.
Among the TSP-specific solvers, the closest to our method at $n = 500$ is LEHD
with random re-construction, at a higher gap and more than $20\times$ our cost,
and at $n = 1000$ it is DRHG with $10\times$ the cost.
At $n = 10000$, \ours{} matches COExpander and attains a lower gap than every
other diffusion solver.
A considerable gap remains between these general neural solvers and the hybrid
methods that run Lin--Kernighan or destroy-and-repair search inside their loops.
These methods reach lower gaps at a lower cost.

\subsection{Scaling test-time compute}
\label{sec:scaling}

\begin{figure}[t]
\begin{center}
\includegraphics[width=\linewidth]{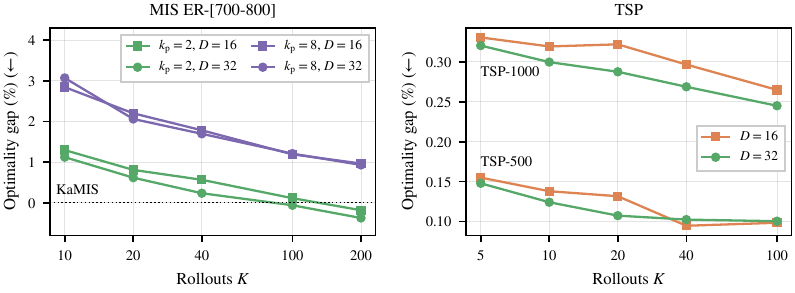}
\end{center}
\caption{Test-time scaling.
Left: MIS on ER-[700-800], optimality gap against the number of rollouts $K$
(log scale) for sequential peeling with $k_{\mathrm{p}} \in \{2, 8\}$ nodes per
round at depth $D \in \{16, 32\}$.
Right: TSP-500 and TSP-1000, gap after 2-opt against $K$ for
$D \in \{16, 32\}$.}
\label{fig:scaling}
\end{figure}

\begin{figure}[t]
\begin{center}
\includegraphics[width=\linewidth]{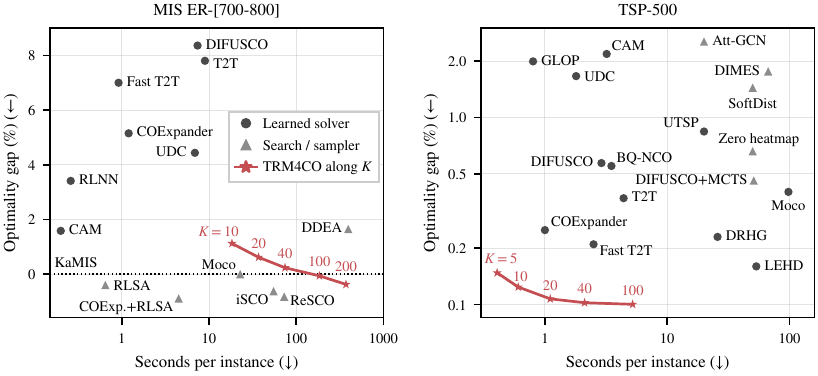}
\end{center}
\caption{Quality against wall-clock time.
We plot optimality gap against seconds per instance on one RTX 4090 for
ER-[700-800] (left) and TSP-500 (right).
The MIS gap is measured against the KaMIS reference (dotted line) of
Table~\ref{tab:mis}, so negative values mean that the reference is beaten.
Circles are learned solvers, triangles are pipelines dominated by search or
sampling (MCTS, samplers, evolution), and stars are \ours{} for increasing $K$
($k_{\mathrm{p}} = 2$ and $D = 32$ on ER-[700-800], $D = 32$ on TSP-500).}
\label{fig:pareto}
\end{figure}

Figure~\ref{fig:scaling} shows the contribution of the two scaling axes.
On ER-[700-800], width matters more than depth.
At $k_{\mathrm{p}} = 8$ the gap falls steadily with $K$, while the curves for
$D \in \{16, 32\}$ lie on top of each other.
Finer peeling helps most.
With $k_{\mathrm{p}} = 2$ the gap is lower at every $K$ and crosses the KaMIS
reference at both depths, since every peel round takes the best of $K$ fresh
rollouts on the residual graph.
Depth adds a small but consistent gain at $k_{\mathrm{p}} = 2$.
On TSP the gap after 2-opt falls with $K$ at both depths, and on TSP-1000 it is
still falling at the largest width we ran.
Overall cost is linear in $K \times D$ because the rollouts run as one batched
forward.
Figure~\ref{fig:pareto} places these points among the baselines.
On TSP-500 the \ours{} curve lies below every learned solver.
On ER-[700-800] it reaches the lowest gap of the learned solvers without a
sampler, but RLSA and COExpander with RLSA reach the same or lower gaps at a
small fraction of the cost, so there the samplers dominate.

\subsection{Ablations}
\label{sec:ablation}

Table~\ref{tab:ablation} in Appendix~\ref{app:ablation} varies one factor at a
time.
First, the structure of a combinatorial optimization problem has to enter the
network in an appropriate way.
For MIS, that is the self-attention.
A message-passing GNN encoder in front of the recursion leaves a one-shot gap up
to twice that of the edge-biased attention from Eq.~(\ref{eq:edgebias}), which
needs no encoder at all.
For TSP, it is the input encoding.
Raw coordinates leave a gap $5.6\times$ that of sinusoidal features at the scale
of nearest-neighbor distances.
However, a $k$-nearest-neighbor~(kNN) edge bias, as on MIS, adds nothing, and a
finer sinusoidal embedding hurts.
Second, decoding composes with the recursion.
On MIS, sequential peeling cuts the gap of the same checkpoint by two thirds,
and more width and depth halve what remains.
On TSP, the greedy walk strands nodes and recovers only half of the loss even
with multi-start, whereas edge insertion cannot strand nodes, so its raw tours
have half the gap of the multi-start walk, and one 2-opt pass brings them within
1\% of optimal.
Third, the prefix tokens provide global memory.
Removing them raises the one-shot MIS gap by half at the same schedule, so a few
registers that every node can attend to give the recursion a scratchpad that
node tokens alone do not.

\subsection{Self-relabeling as self-supervision}
\label{sec:selfsup}

\begin{table}[!t]
\caption{Self-relabeling.
Test gap in \% for MIS on ER-[700-800] against KaMIS, decoded with sequential
peeling at the setting of Table~\ref{tab:mis} ($k_{\mathrm{p}} = 2$,
$\tau = 0.5$, $K = 10$, $D = 32$), and for TSP-500 against Concorde at the
setting of Table~\ref{tab:tsp} ($K = 40$, $D = 32$, 2-opt).
The supervised rows train on solver labels, the continuation row trains on top
of the supervised checkpoint, and the from-scratch rows start from random
initialization without any solver labels.}
\label{tab:selfsup_main}
\centering
\small
\begin{tabular}{llr}
\toprule
Run & Labels & Gap \% \\
\midrule
\multicolumn{3}{l}{\emph{MIS ER-[700-800]}} \\
Supervised & KaMIS & 2.53 \\
Self-relabel continuation & KaMIS, then own & \textbf{1.11} \\
Self-relabel from scratch & Own only & 1.43 \\
\midrule
\multicolumn{3}{l}{\emph{TSP-500}} \\
Supervised & LKH-3 & \textbf{0.11} \\
Self-relabel from scratch & Own only & 2.26 \\
\bottomrule
\end{tabular}
\end{table}

We can generate and improve the labels throughout training by keeping the best
solution (Section~\ref{sec:algorithm}).
Table~\ref{tab:selfsup_main} compares this with supervised training on
ER-[700-800] and TSP-500, under the same decoding settings as
Tables~\ref{tab:mis} and~\ref{tab:tsp}.

\paragraph{Labels beyond the solver.}
Continuing the supervised MIS checkpoint with the labels replaced by its own
better solutions cuts the test gap by more than half and gives our best
checkpoint, the one behind Table~\ref{tab:mis}.
This indicates that self-relabeling can further improve a trained model, since
it then trains on larger sets than those found by the solver for the initial
training labels.
The relabeling passes raise about 16\% of the training instances above their
KaMIS labels, by one node on average.

\paragraph{No solver at all.}
Starting from random initialization with no labels available, the model first
seeds the initial labels with its own solution guesses and then hill-climbs on
them.
Once there is a signal, the model exploits it in training and keeps
self-relabeling.
On MIS its labels end up within one node of the KaMIS labels on average and
match or beat them on 42\% of the training graphs.
Its test gap comes within one third of a point of the supervised continuation
and is below that of the supervised checkpoint itself.
On TSP-500 the label-free run closes most of the distance to the LKH-3 tours,
from about 10\% above them to less than 2\%, but its test gap is still
$20\times$ that of the supervised model.
Figure~\ref{fig:selfsup} in Appendix~\ref{app:selfsup} traces the evolution of
the training labels.

\section{Related work}
\label{sec:related}

\paragraph{Constructive policies.}
Autoregressive policies build a solution one decision at a time
\citep{vinyals2015pointer,kool2019attention,kwon2020pomo}.
LEHD \citep{luo2023lehd} and BQ-NCO \citep{drakulic2023bqnco} train on small
instances and then generalize.
GLOP \citep{ye2024glop}, UDC \citep{zheng2024udc}, DualOpt
\citep{zhou2025dualopt}, and H-TSP \citep{pan2023htsp} reach larger instances by
decomposition.
HyperNS \citep{lu2026hyperns} instead guides a Lin--Kernighan search with a
heatmap, and GOAL \citep{drakulic2025goal} injects edges into attention as we
do.

\paragraph{Heatmap solvers and search.}
Heatmap solvers score all variables in one pass and leave construction to a
search.
Att-GCN \citep{fu2021attgcn} pairs a learned heatmap with MCTS for large TSP
instances, and DIMES \citep{qiu2022dimes} makes the heatmap a differentiable
policy.
DIFUSCO \citep{sun2023difusco} models the heatmap with a diffusion model, and
T2T \citep{li2023t2t}, Fast T2T \citep{li2024fastt2t}, DISCO
\citep{zhao2025disco}, CADO \citep{song2026cado}, DDEA
\citep{salvasoler2026ddea}, and PCI \citep{basson2026pci} refine its training,
sampling, or search.
\ours{} instead refines its prediction by recursion inside the network.
\citet{xia2024rethinking} and \citet{pan2026beyondheatmap} show that the search
dominates the heatmap in these pipelines, and \citet{boether2022whatswrong} and
\citet{angelini2023modern} reach a similar verdict for MIS.
FrontierCO \citep{feng2026frontierco} and the survey of \citet{ba2026survey}
find neural solvers behind tuned heuristics at scale.

\paragraph{Unsupervised and self-improving solvers.}
UTSP \citep{min2023utsp}, DiffUCO \citep{sanokowski2024diffuco}, SDDS
\citep{sanokowski2025sdds}, and CAM \citep{feng2026cam} train without solver
labels by optimizing relaxed objectives.
Moco \citep{dernedde2025moco} learns a heatmap optimizer that spends its compute
at test time.
SIL \citep{luo2024sil} improves a constructive policy on its own solutions after
local search.
Samplers such as iSCO \citep{sun2023isco} and ReSCO \citep{li2025resco} need no
learning at all.
Our self-relabeling is closest to SIL, with the recursive model's own decoded
solutions as labels and no relaxed objective.

\paragraph{Recursive reasoning.}
HRM \citep{wang2025hrm} and TRM \citep{jolicoeur2025trm,sghaier2026ptrm} solve
puzzles by latent recursion with adaptive halting.
Looped Transformers reuse one block of layers to gain depth
\citep{wang2026smelt,suleymanzade2026loopedflows}.
To the best of our knowledge, we are the first to apply latent recursion to
combinatorial optimization at this scale.

\section{Conclusion}
\label{sec:conclusion}
We have presented \ours{}, a small recursive network with problem-specific
encoding and decoding that is competitive with or better than search-heavy
pipelines on MIS and TSP.
Its key ingredient is test-time scaling in depth and in width on the same
weights, which efficiently turns compute into solution quality.
Self-relabeling improves the model further, on top of solver labels or without
any.
\ours{} attacks the remaining gap between neural solvers and learning-free
search from the perspective of recursive reasoning.
Our backbone is general, but the encoder and decoder remain problem-specific,
and reducing that dependence is the natural next step.


\subsection*{Reproducibility statement}
The code and checkpoints of every reported run, and the relabeled datasets will
be released upon acceptance.
Appendix~\ref{app:hyper} lists the hyperparameters, Appendix~\ref{app:data} the
datasets and reference solutions, and Appendix~\ref{app:protocol} the runtime
protocol.

\bibliography{iclr2027_conference}
\bibliographystyle{iclr2027_conference}

\clearpage
\appendix
\raggedbottom \makeatletter\setlength{\@fptop}{0pt}\makeatother  
\section{Algorithm and problem-specific hooks}
\label{app:alg}

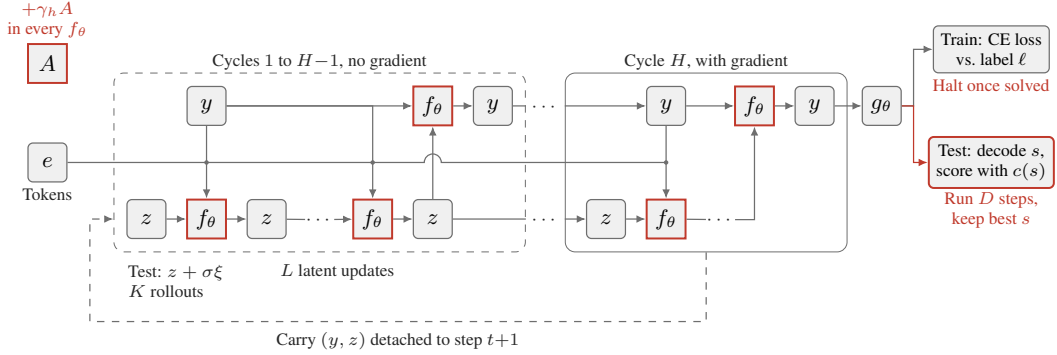
\begin{figure}[!htb]
\begin{center}
\resizebox{\textwidth}{!}{\input{figures/recursion}}
\end{center}
\caption{One recursion step of \ours{}, unrolled.
Gray is the TRM schedule of Eq.~(\ref{eq:trm}), with $L$ latent updates and one
answer update per cycle, $H$ cycles per step, gradient only through the last
cycle, and $(y, z)$ carried detached to the next step.
Test-time noise on $z$ \citep{sghaier2026ptrm} gives $K$ independent rollouts.
Red is what we change.
For MIS, the adjacency $A$ biases every attention call (Eq.~\ref{eq:edgebias}).
At test time, every step (TSP) or every peeling round (MIS) is decoded into a
feasible solution and scored with the objective.
In training, the solved check of Table~\ref{tab:hooks} halts an instance in
place of the learned halting signal.}
\label{fig:recursion}
\end{figure}

Algorithm~\ref{alg:trm4co} lists the procedures of Section~\ref{sec:algorithm}
that MIS and TSP share, and Table~\ref{tab:hooks} the problem-specific hooks it
calls.
Figure~\ref{fig:recursion} unrolls one recursion step.

\begin{table}[!htb]
\caption{Problem-specific hooks of Algorithm~\ref{alg:trm4co}.
Everything not listed here (the core, the recursion schedule, deep supervision,
and the self-relabeling loop) is shared.}
\label{tab:hooks}
\begin{center}
\small
\setlength{\tabcolsep}{4pt}
\begin{tabular}{lll}
\toprule
Hook & MIS & TSP \\
\midrule
Instance $x$ & Adjacency $A \in \{0,1\}^{n \times n}$ & Coordinates $X \in [0,1]^{n \times 2}$ \\
Tokens $\mathrm{Emb}(x)$ & Linear$[\deg,\ \mathrm{RWPE},\ u \sim \mathcal{U}(0,1)]$ & Linear$[X,\ \sin/\cos(2^{l}\pi X)]_{l < L_{\mathrm{pe}}}$ \\
Attention bias & Adjacency $A$ & None (standard self-attention) \\
Head $g_\theta(y)$ & Node logits $[n, 2]$ & Bilinear successor logits $[n, n]$ \\
Decoder $\mathrm{Dec}$ & Greedy independent set in score order & Greedy edge insertion by $\bar g_{ij}/d_{ij}$ \\
Wrapper $\mathcal{W}$ & Sequential peeling around the recursion & 2-opt on the tour returned by \textsc{Solve} \\
Objective $c(s)$ & $|S|$ (max) & Tour length (min) \\
$\mathrm{Solved}(g_\theta(y), \ell, x)$ & Greedy $|S| \geq |\ell|$ & Successors match the canonical $\ell$ \\
Labels $\ell$ & KaMIS 10\,s (ER), exact (SATLIB) & LKH-3 (1000 trials) \\
Augmentation & Random peel of label nodes & Dihedral-8 of the unit square \\
\bottomrule
\end{tabular}
\end{center}
\end{table}

\begin{algorithm}[!htb]
\caption{\ours{}: one recursion step, inference with depth $D$ and width $K$,
and training.
Problem-specific hooks are listed in Table~\ref{tab:hooks}, and the rest is
shared between MIS and TSP.}
\label{alg:trm4co}
\begin{algorithmic}[1]
\Require core $f_\theta$, head $g_\theta$, prefix tokens $p$, initial states $y_0, z_0$
\Require hooks $\mathrm{Emb}$, $\mathrm{Dec}$, $c$, $\mathrm{Solved}$, and wrapper $\mathcal{W}$ of Table~\ref{tab:hooks}
\Procedure{Step}{$y, z, e, A$} \Comment{Eq.~(\ref{eq:trm}), with Eq.~(\ref{eq:edgebias}) inside $f_\theta$}
  \For{$i = 1, \dots, H$} \Comment{cycles $1 \dots H{-}1$ without gradient}
    \For{$l = 1, \dots, L$}
      \State $z \leftarrow f_\theta(z + y + e;\ A)$
    \EndFor
    \State $y \leftarrow f_\theta(z + y;\ A)$
  \EndFor
  \State \Return $y, z$
\EndProcedure
\Statex
\Procedure{Solve}{$x$; $D, K, \sigma$} \Comment{inference; $x$ may be a batch of instances}
  \State $e \leftarrow [p;\ \mathrm{Emb}(x)]$
  \State $A \leftarrow$ adjacency of $x$ (MIS) or $0$ (TSP)
  \State $(y^{(j)}, z^{(j)}) \leftarrow (y_0, z_0)$ for $j = 1, \dots, K$
  \State $s^\star \leftarrow$ none
  \For{$t = 1, \dots, D$} \Comment{depth}
    \For{$j = 1, \dots, K$ \textbf{in one batched forward pass}} \Comment{width}
      \State $z^{(j)} \leftarrow z^{(j)} + \sigma \xi$ with $\xi \sim \mathcal{N}(0, I)$
      \State $(y^{(j)}, z^{(j)}) \leftarrow$ \Call{Step}{$y^{(j)}, z^{(j)}, e, A$}
      \State $s \leftarrow \mathrm{Dec}(g_\theta(y^{(j)}), x)$
      \If{$s^\star$ is none or $c(s)$ improves on $c(s^\star)$}
        \State $s^\star \leftarrow s$
      \EndIf
    \EndFor
  \EndFor
  \State \Return $s^\star$
\EndProcedure
\Statex
\Procedure{Train}{$\{(x_i, \ell_i)\}$; $D, T$} \Comment{deep supervision with problem-specific halting}
  \State fill $B$ batch slots with augmented instances, each with $e$, $A$, $(y_0, z_0)$, and $t = 0$
  \Loop
    \State $(y, z) \leftarrow$ \Call{Step}{$y, z, e, A$} on all slots
    \State $t \leftarrow t + 1$
    \State $\mathcal{L} \leftarrow \mathrm{CE}(g_\theta(y), \ell)$
    \State update $\theta$ with $\nabla_\theta \mathcal{L}$
    \State detach $(y, z)$
    \For{each slot with $\mathrm{Solved}(g_\theta(y), \ell, x)$ or $t = D$}
      \State refill the slot with a new augmented instance
    \EndFor
    \If{$T$ epochs have passed since the last relabel pass}
      \State $\ell_i \leftarrow$ the better of $\ell_i$ and $\mathcal{W}(x_i)$\quad for all $i$ \Comment{self-relabel}
    \EndIf
  \EndLoop
\EndProcedure
\end{algorithmic}
\end{algorithm}

\section{Hyperparameters}

\label{app:hyper}

\paragraph{Training.}
The loss is a cross-entropy per node (MIS) or per successor (TSP) against the
labels of Table~\ref{tab:hooks}, with peel augmentation for MIS and the eight
symmetries of the square for TSP applied on the fly (see below).
Each model trains on two H100 GPUs.
Several reported checkpoints continue earlier runs, and the epoch counts in
Table~\ref{tab:hyper} are summed over all stages.
The ER supervised checkpoint runs 10,000 epochs with the edge-biased
architecture, 5000 at $D = 64$ and 5000 at $D = 32$, warm-started from a run
with a message-passing encoder.
Its self-relabeling continuation adds 3000 epochs.
The SATLIB and TSP-10000 checkpoints continue a first run of 1000 epochs for
another 1000.
The label-free ER chain of Table~\ref{tab:selfsup_main} runs 5000 epochs with
one-shot relabeling and three stages of 3000 epochs with peeling, and the
label-free TSP-500 chain runs stages of 500, 500, and 1000 epochs.

\paragraph{Implementation details.}
The halting head of TRM is retained, but we block its gradient.
Its loss is still computed with weight $0.5$ as in TRM for logging, and the head
is used neither for halting nor for selection.
As in TRM, a random minimum step count is imposed on a small fraction of
training instances so that later steps keep receiving a gradient.
The loss is the stablemax cross-entropy of \citet{jolicoeur2025trm}.
Body weights are trained with Muon \citep{jordan2024muon}, heads and biases with
AdamW, and prefix tokens with sign-SGD.
An exponential moving average of the weights is maintained.
Training uses $D = 16$ for TSP and $32$ for MIS, $P = 16$ prefix tokens, $H = 3$
cycles of $L = 6$ latent updates, and a 12-step random-walk positional encoding
for MIS (Table~\ref{tab:hyper}).
The TSP solved check compares predicted successors with the canonical
(counterclockwise) label because decoding a tour would cost $n$ sequential steps
per training step.
The 2-opt repair may be restricted to the $m$ longest edges of the tour.
The main-table numbers use a full scan.
During self-relabeling, the TSP tours of a relabel pass are repaired with 2-opt
and canonicalized before the comparison with the stored label.

\begin{table}[!htb]
\caption{Hyperparameters of the reported models.
Epochs are passes over the training set, summed over all stages of a continued
run.
The self-relabeling continuation of Table~\ref{tab:selfsup_main} adds 3000
epochs with a 2000-step warm-up.
Test-time $K$, $D$, $\sigma$, $\tau$, $k_{\mathrm{p}}$, and $m$ are given with
each table.}
\label{tab:hyper}
\begin{center}
\footnotesize
\setlength{\tabcolsep}{3pt}
\begin{tabular}{ll}
\toprule
Parameter & Value \\
\midrule
Hidden size, heads, blocks & 512, 8, 2 \\
MLP expansion & 4 \\
Recursion $H$, $L$ & 3, 6 \\
Prefix tokens $P$ & 16 \\
Structure, MIS & Edge-biased attention, 12-step random-walk positional encoding \\
Structure, TSP & Self-attention, $L_{\mathrm{pe}}=6$ coordinate encoding \\
Successor head dimension $d_s$ & 128 (TSP) \\
Training depth $D$ & 32 (MIS; 64 in the first ER stage), 16 (TSP) \\
Global batch size & 768 (ER, TSP-500), 384 (SATLIB), 512 (TSP-1000), 48 (TSP-10000) \\
Epochs, MIS & 10,000 (ER), 2000 (SATLIB) \\
Epochs, TSP & 500 (TSP-500), 1000 (TSP-1000), 2000 (TSP-10000) \\
Learning rate & AdamW $10^{-4}$, Muon $2 \cdot 10^{-3}$, 5000 warm-up steps, cosine to 1\% \\
Weight decay, EMA rate & 0.1, 0.999 \\
Augmentation & Peel, prob.\ $0.9$, depth $\leq 9$ (ER), $\leq 32$ (SATLIB); dihedral-8 (TSP) \\
Self-relabel pass, ER continuation & Every 200 epochs, $K = 4$, $D = 16$, $k_{\mathrm{p}} = 2$ \\
\bottomrule
\end{tabular}
\end{center}
\end{table}

\paragraph{Augmentations.}
Two augmentations run on the fly.
For MIS, peel augmentation removes a random number of label nodes one at a time,
each with its closed neighborhood, so that training also sees the residual
graphs that sequential peeling visits.
For TSP, one of the eight symmetries of the square is applied with the tour
orientation kept canonical (counterclockwise).

\section{Datasets and references}
\label{app:data}

\paragraph{MIS.}
The ER-[700-800] and SATLIB graphs are those distributed with DIMES
\citep{qiu2022dimes}.
We label the ER training graphs with KaMIS \citep{lamm2017kamis} at a 10-second
budget per graph, which gives a mean set size of 44.52 on the training graphs
and 44.63 on the test graphs.
DIFUSCO and the papers that follow it compare against a longer KaMIS run on the
test graphs with a mean size of 44.87 (24 seconds per graph), which
Table~\ref{tab:mis} and Figures~\ref{fig:scaling} and~\ref{fig:pareto} use as
the reference.
The ablation table (Table~\ref{tab:ablation}) reports gaps against our own test
labels, 44.63, so its gaps are about half a point lower for the same set size.
SATLIB labels are exact, computed with Gurobi \citep{gurobi2026}.
Their mean of 425.96 on the test graphs coincides with the KaMIS reference of
those papers.

\paragraph{TSP.}
Training instances are drawn uniformly from the unit square and labeled with
LKH-3 \citep{helsgaun2017lkh3} using 1000 trials and the best of 10 runs per
instance.
The test sets are the 128, 128, and 16 instances of \citet{fu2021attgcn}.
The reference tours that ship with their public copies are not optimal.
The tours attributed to Concorde \citep{applegate2006concorde} are 0.23\% and
0.47\% above the optimum at $n = 500$ and $1000$, and the $n = 10000$ file
carries placeholder tours.
We therefore recompute the references, with Concorde at $n = 500$ and $1000$
(certified optimal, mean lengths 16.5458 and 23.1181) and with LKH-3 at
$n = 10000$ (10,000 trials, best of 10 runs, mean length 71.7708, the LKH-3
value quoted in the literature).
All gaps in Table~\ref{tab:tsp} are against these references, and published
lengths are re-expressed accordingly (Appendix~\ref{app:tablenotes}).

\section{Runtime protocol}
\label{app:protocol}
All runtimes in Tables~\ref{tab:mis} and~\ref{tab:tsp} are our own measurements
on one RTX 4090 with the released code and checkpoints of each method, in the
inference setting whose quality number the table quotes.
For each compared method, we time the complete pipeline from the instance to the
final solution, including heatmap generation, decoding, and any post-processing
such as 2-opt, rewriting, or search, on a subset of our test instances (8, 4,
and 2 instances at $n = 500$, $1000$, and $10000$ for TSP, and 8 graphs for
MIS), and report the mean per instance after a warm-up instance or batch.
Batched solvers (DRHG, SIL, LEHD, BQ-NCO, iSCO, ReSCO) are timed on batches of 2
to 8 instances, and the batch time is divided by the batch size, which
overstates their per-instance cost relative to the batch sizes in their papers.
MCTS pipelines run the search of \citet{fu2021attgcn} at the time limit of the
respective codebase.
The limit is 0.1 seconds per city for DIFUSCO, DIMES, SoftDist, and the controls
of \citet{pan2026beyondheatmap}, and 0.04 seconds per city for Att-GCN and UTSP,
so their runtime is set by that budget plus the heatmap time.
Methods without public code (DISCO, CADO, HyCO, PCI, HyperNS) or without
released weights for our benchmarks (DIMES on MIS, GFlowNets, LwD, DiffUCO,
SDDS) have no runtime, as do settings that exceed the 24 GB memory of the card
(T2T and Fast T2T sampling at $n = 10000$).
Our own TSP-10000 row exceeds the memory of the card as well, since its 40
rollouts do not fit in 24 GB at once.
We run it on an RTX 5090 (340 seconds per instance) and scale the time by the
4090-to-5090 runtime ratio of TSP-1000 at the same $K$ and $D$ (4.43 to 4.13
seconds), which gives the 364 seconds in Table~\ref{tab:tsp} with the unrounded
timings.
Our own runtimes use the full test sets in batches of 320 rollouts (for example,
32 instances at $K = 10$ and 8 at $K = 40$) and exclude the first batch, which
includes compilation.
The MIS $K = 200$ point uses 6 graphs in batches of 2.
Table~\ref{tab:cost} lists the parameter counts of the inference networks.

\begin{table}[!t]
\caption{Learned parameters of each network.
Ours and Moco are counted from checkpoints; the others are derived from stated
hyperparameters or official code, since no paper states its count.}
\label{tab:cost}
\begin{center}
\small
\begin{tabular}{lrr}
\toprule
Method & MIS & TSP \\
\midrule
Moco & 0.36M & 0.85M \\
T2T / Fast T2T & 1.34M & 5.33M \\
DIFUSCO / DISCO & -- & 5.33M \\
CAM & 0.18M & 5.20M \\
UDC & 4.35M & 1.56M \\
DRHG & -- & 2.65M \\
HyperNS (heatmap network only) & -- & $\sim$11.1M \\
\ours{} & 6.83M & 6.96M \\
\bottomrule
\end{tabular}
\end{center}
\end{table}

\section{Notes on the main tables}
\label{app:tablenotes}

\paragraph{Sources of the quality numbers.}
Every quality number of a compared method is the one reported in its own paper,
in its main sampling setting.
Where a paper does not cover one of our benchmarks but a compared paper
evaluates the method on it, we use that number.
DiffUCO and SDDS on ER-[700-800] are the re-implementations of
\citet{feng2026cam}, LwD's ER number is the rerun of \citet{qiu2022dimes}, and
T2T and Fast T2T at $n = 10000$ for TSP are evaluated by \citet{zhao2025disco}
and \citet{ma2025coexpander}.
Zero heatmap and GT-Prior are the heatmap-free controls of
\citet{pan2026beyondheatmap}.
For TSP, where a paper prints only the length or only the gap, the other is
derived against our reference, and where it uses another reference, the gap is
recomputed against ours.

\paragraph{Own instance sets.}
The dagger marks methods evaluated on their own generated instances of the same
distribution, with the gap against their own reference run.
Where a method's code and checkpoints are released, we re-evaluate it on our
test sets in the setting of its paper row and report that number without a
dagger.
These are COExpander with and without RLSA on both MIS benchmarks, and UDC,
BQ-NCO, DRHG, and LEHD on TSP.
Their full-set results agree with their own-instance numbers to within 0.1
points, except for COExpander with RLSA (45.27 against 44.98 on ER-[700-800] and
425.45 against 425.32 on SATLIB) and DRHG at $n = 500$ (0.23\% against the
reported 0.11\%, with either released checkpoint).
The remaining daggers are HyCO and GFlowNets on both MIS benchmarks, LwD on
SATLIB, and PCI at $n = 500$ and $1000$, whose code or checkpoints are not
released.
Lengths and sizes on different instances are not directly comparable, so the gap
is the number to compare.

\paragraph{Type and Search columns.}
In Table~\ref{tab:mis}, S + RL combines supervised and reinforcement learning,
as supervised pretraining followed by RL fine-tuning for CADO and as an RL
constructor whose solution prefix is completed by a supervised diffusion model
for HyCO.
S + self is supervised training followed by self-relabeling (ours), and the
suffixes ``+ search'', ``+ sampler'', and ``+ evolution'' name the
objective-guided gradient search of T2T and Fast T2T, the RLSA sampler applied
after COExpander, and the evolutionary loop of DDEA.
U + search is Moco's meta-optimizer loop.
In Table~\ref{tab:tsp}, MCTS denotes the search of \citet{fu2021attgcn} at the
time limit of the respective codebase (Appendix~\ref{app:protocol}), and RRC,
PRC, revisers, and conquer are the iterative neural reconstruction loops of
LEHD, SIL, GLOP, and UDC.

\paragraph{Reference solvers.}
KaMIS, Concorde, and LKH-3 are listed with the CPU times reported by
\citet{qiu2022dimes} and \citet{fu2021attgcn}, divided by the number of
instances, and are not comparable with the GPU times of the other rows.

\section{Full ablations}
\label{app:ablation}

Table~\ref{tab:ablation} lists every variant behind the findings of
Section~\ref{sec:ablation}.

\begin{table}[!htb]
\caption{Ablations, one factor at a time.
MIS on ER-[700-800] (gap in \% against our KaMIS labels, 44.63), TSP-500 (gap in
\% against Concorde), and TSP-10000 (gap in \% against LKH-3).
Each block varies one factor of the default configuration, with $K$, $D$, and
$\sigma$ as noted.
The MIS decoding block uses the supervised checkpoint of
Table~\ref{tab:selfsup_main} with $\sigma = 0.5$ and $\tau = 0$, and the rollout
and Gumbel rows an earlier checkpoint with $\sigma = 0.7$, so their gaps are not
directly comparable with Table~\ref{tab:mis}.}
\label{tab:ablation}
\begin{center}
\small
\begin{tabular}{llr}
\toprule
Factor & Variant & Gap \% \\
\midrule
\multicolumn{3}{l}{\emph{MIS structure injection (one-shot, $K{=}1$)}} \\
 & 3-layer GNN encoder + self-attention & 20.2 \\
 & 6-layer GNN encoder + self-attention & 13.5 \\
 & No encoder + edge-biased attention (default) & 10.0 \\
\multicolumn{3}{l}{\emph{MIS prefix tokens (same schedule, best checkpoint of the last training window)}} \\
 & $P = 0$ & 15.0 \\
 & $P = 16$ (default) & 10.0 \\
\multicolumn{3}{l}{\emph{MIS decoding (same checkpoint, $K{=}10$, $D{=}32$)}} \\
 & One-shot greedy & 9.06 \\
 & Sequential peel, $k_{\mathrm{p}}{=}2$ & 2.67 \\
 & Sequential peel, $k_{\mathrm{p}}{=}2$, $K{=}40$, $D{=}64$ & 1.14 \\
\multicolumn{3}{l}{\emph{MIS Gumbel noise on peeling ($k_{\mathrm{p}}{=}2$, $\sigma{=}0.7$, $K{=}4$)}} \\
 & Sequential peel, $\tau{=}0$ & 6.61 \\
 & Sequential peel, $\tau{=}0.1$ & 6.42 \\
 & Sequential peel, $\tau{=}0.3$ & 5.86 \\
 & Sequential peel, $\tau{=}1.0$ & 6.96 \\
\multicolumn{3}{l}{\emph{MIS rollouts ($k_{\mathrm{p}}{=}2$, $\sigma{=}0.7$)}} \\
 & $K{=}4$ & 7.0 \\
 & $K{=}100$ & 1.97 \\
\midrule
\multicolumn{3}{l}{\emph{TSP-500 input encoding (same schedule, best in-run checkpoint)}} \\
 & Raw coordinates, self-attention & 15.89 \\
 & Raw coordinates, kNN edge-biased attention & 17.27 \\
 & $L_{\mathrm{pe}}{=}6$, self-attention (default) & 2.84 \\
 & $L_{\mathrm{pe}}{=}6$, kNN edge-biased attention & 3.20 \\
 & $L_{\mathrm{pe}}{=}12$, self-attention & 6.72 \\
\multicolumn{3}{l}{\emph{TSP-500 decoding (same checkpoint, $K{=}10$, $D{=}16$)}} \\
 & Walk, one-shot ($K{=}1$, last step) & 8.44 \\
 & Walk, argmax & 2.76 \\
 & Walk, multi-start & 1.36 \\
 & Walk, multi-start + 2-opt ($m{=}16$) & 0.247 \\
 & Walk, multi-start + $\tau{=}0.3$ + full 2-opt & 0.169 \\
 & Insertion, raw & 0.611 \\
 & Insertion + full 2-opt (default) & 0.151 \\
\multicolumn{3}{l}{\emph{TSP-10000 2-opt budget $m$ (insertion, $K{=}4$, $D{=}16$)}} \\
 & $m{=}16$ & 2.73 \\
 & $m{=}64$ & 2.19 \\
 & $m{=}256$ & 1.91 \\
 & $m{=}1024$ & 1.77 \\
 & Full scan & 1.69 \\
\bottomrule
\end{tabular}
\end{center}
\end{table}

\section{Self-relabeling details}
\label{app:selfsup}

\begin{figure}[!htb]
\begin{center}
\includegraphics[width=\linewidth]{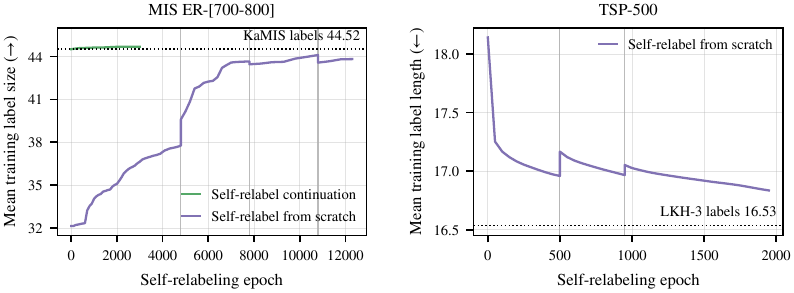}
\end{center}
\caption{Training labels under self-relabeling, for the runs of
Table~\ref{tab:selfsup_main}.
Mean size (MIS, left) or length (TSP-500, right) of the training labels against
the epochs of self-relabeling, for the continuation of the supervised MIS
checkpoint and for the label-free chains from random initialization.
Vertical lines mark the warm-started stages of a chain.
Each stage restarts its labels from the model's own first pass, which produces
the steps in the curves.
The dotted line is the mean of the solver labels.}
\label{fig:selfsup}
\end{figure}

Figure~\ref{fig:selfsup} traces the training labels of the runs of
Table~\ref{tab:selfsup_main} over the epochs of self-relabeling.


\paragraph{A decoder mismatch pitfall.}
Relabeling with one-shot decoding produces the best one-shot model among all our
checkpoints, but it inverts the benefit of sequential peeling.
At the same training stage, the supervised model still gains 3 to 5 points from
peeling, while the relabeled model loses 1 to 3.
Our conjecture is that training toward the model's own argmax sets makes scores
mean ``in my preferred set'' rather than ``safe in any near-optimal set'', so
top-score commitments strand later rounds.
Relabeling and selecting checkpoints with the deployment decoder removes the
inversion and gives our best MIS checkpoint, the one behind the rows of
Table~\ref{tab:mis}.

\end{document}

%% file: math_commands.tex
\usepackage{amsmath,amsfonts,bm}

\def\eqref#1{equation~\ref{#1}}

\def\1{\bm{1}}

\DeclareMathAlphabet{\mathsfit}{\encodingdefault}{\sfdefault}{m}{sl}
\SetMathAlphabet{\mathsfit}{bold}{\encodingdefault}{\sfdefault}{bx}{n}



%% file: figures/framework.tex
\begin{tikzpicture}[
  x=1cm, y=1cm, font=\footnotesize, >=Latex, line width=0.5pt,
  panel/.style={draw=gray!50, rounded corners=6pt, inner sep=0pt, fill=gray!4},
  ptitle/.style={font=\footnotesize\bfseries, anchor=north, inner sep=4pt},
  box/.style={draw=gray!55!black, fill=gray!12, rounded corners=2pt, align=center,
              text width=2.4cm, minimum height=5mm, inner sep=2pt, line width=0.4pt},
  sbox/.style={box, text width=1.0cm},
  ebox/.style={box, draw=blue!55!black, fill=blue!16},
  dbox/.style={box, draw=green!45!black, fill=green!16},
  ours/.style={draw=ourscol, line width=0.8pt},
  note/.style={font=\scriptsize, align=center, inner sep=1pt, text=gray!25!black},
  onote/.style={note, text=ourscol},
  arr/.style={->, gray!30!black},
  oarr/.style={->, ourscol},
  darr/.style={->, dashed, ourscol},
  zloop/.style={->, blue!55!black},
  yloop/.style={->, orange!75!black},
  vtx/.style={circle, draw=gray!70!black, fill=white, inner sep=0pt, minimum size=4pt},
  vsel/.style={vtx, fill=ourscol, draw=ourscol},
  city/.style={circle, fill=gray!70!black, inner sep=0pt, minimum size=2.6pt},
]

\newcommand{\misgraph}[3]{
  \begin{scope}[shift={(#1,#2)}]
    \coordinate (a) at (0.00,0.90); \coordinate (b) at (0.50,1.00); \coordinate (c) at (1.00,0.85);
    \coordinate (d) at (0.15,0.40); \coordinate (e) at (0.60,0.50); \coordinate (f) at (1.05,0.35);
    \coordinate (g) at (0.35,0.00); \coordinate (h) at (0.85,0.05);
    \foreach \u/\v in {a/b,b/c,a/d,b/e,c/f,d/e,e/f,d/g,e/g,e/h,f/h,g/h,b/d}
      \draw[gray!55] (\u) -- (\v);
    \foreach \p in {b,c,d,e,h} \node[vtx] at (\p) {};
    \ifnum#3=1 \foreach \p in {a,f,g} \node[vsel] at (\p) {};
    \else \foreach \p in {a,f,g} \node[vtx] at (\p) {}; \fi
  \end{scope}}
\newcommand{\tspinst}[3]{
  \begin{scope}[shift={(#1,#2)}]
    \draw[gray!45] (0,0) rectangle (1,1);
    \coordinate (p1) at (0.10,0.15); \coordinate (p2) at (0.45,0.05); \coordinate (p3) at (0.90,0.20);
    \coordinate (p4) at (0.95,0.60); \coordinate (p5) at (0.70,0.90); \coordinate (p6) at (0.35,0.95);
    \coordinate (p7) at (0.10,0.60); \coordinate (p8) at (0.50,0.50);
    \ifnum#3=1 \draw[ourscol, line width=0.8pt] (p1) -- (p2) -- (p3) -- (p4) -- (p5) -- (p6) -- (p7) -- (p8) -- cycle; \fi
    \foreach \p in {p1,p2,p3,p4,p5,p6,p7,p8} \node[city] at (\p) {};
  \end{scope}}

\node[ptitle] at (0.6, 6.15) {Instance};
\misgraph{0.1}{4.45}{0}
\node[note] at (0.6, 4.15) {MIS: graph};
\tspinst{0.1}{2.1}{0}
\node[note] at (0.6, 1.8) {TSP: cities};

\node[panel, fill=blue!5, minimum width=2.8cm, minimum height=4.7cm, anchor=south] (enc) at (2.95, 1.45) {};
\node[ptitle] at (enc.north) {Encoder};
\node[ebox, ours] (Ebox) at (2.95, 5.05) {Structure\\[1pt] \scriptsize MIS: adjacency};
\node[ebox] (feat) at (2.95, 3.625) {Node features\\[1.5pt] \scriptsize MIS: degree, RWPE, noise\\[-1pt] \scriptsize TSP: sinusoidal};
\node[ebox] (lin)  at (2.95, 2.2) {Linear\\[1pt] \scriptsize $+\,P$ prefix tokens};
\draw[arr] (feat) -- (lin);
\draw[oarr] (1.2, 5.05) -- (Ebox.west);
\draw[gray!30!black] (1.2, 4.6) -- (1.45, 4.6) -- (1.45, 2.6) -- (1.2, 2.6);
\draw[arr] (1.45, 3.625) -- (feat.west);

\begin{scope}[on background layer]  
  \node[panel, fill=gray!14, minimum width=4.7cm, minimum height=6.8cm, anchor=south] at (7.34, 0.44) {};
  \node[panel, fill=gray!8, minimum width=4.7cm, minimum height=6.8cm, anchor=south] at (7.22, 0.32) {};
\end{scope}
\node[panel, fill=white, minimum width=4.7cm, minimum height=6.8cm, anchor=south] (bb) at (7.1, 0.2) {};
\node[ptitle] at (bb.north) {Recursive backbone};
\node[note, anchor=north] at (7.1, 6.5) {Weight-tied, 7M parameters\\ Depth: $D$ steps \quad Width: $K$ rollouts};

\node[sbox, draw=blue!55!black, fill=blue!16] (xin) at (5.7, 1.45) {$e$\\[-1.5pt] \scriptsize tokens};
\node[sbox, draw=orange!60!black, fill=orange!22] (yb)  at (7.0, 1.45) {$y$\\[-1.5pt] \scriptsize answer};
\node[sbox, fill=gray!15] (zb)  at (8.3, 1.45) {$z$\\[-1.5pt] \scriptsize latent};
\node[circle, draw=gray!30!black, inner sep=0pt, minimum size=3mm] (sum) at (7.0, 2.35) {\scriptsize$+$};
\draw[arr] (xin.north) |- (sum.west);
\draw[arr] (yb.north) -- (sum.south);
\draw[arr] (zb.north) |- (sum.east);

\node[draw=gray!45, fill=gray!4, rounded corners=3pt, minimum width=2.75cm, minimum height=2.065cm] (stack) at (7.0, 3.8675) {};
\node[box, fill=orange!22, ours] (sa) at (7.0, 3.41) {\rule{0pt}{8pt}Self-attention\\[1pt] \scriptsize logits $+\,\gamma_h A$};
\node[box, draw=green!45!black, fill=green!16] (mlp) at (7.0, 4.49) {MLP};
\node[note, anchor=east, font=\footnotesize] at (stack.west) {$2\times$\,};
\draw[arr] (sum) -- (stack.south);
\draw[arr] (sa) -- (mlp);
\draw[oarr] (Ebox.east) -- (5.0, 5.05) -- (5.0, 3.41) -- (sa.west);
\draw[arr] (lin.east) -- (4.52, 2.2) -- (4.52, 1.45) -- (xin.west);

\draw[gray!30!black] (stack.north) -- (7.0, 5.55);
\draw[zloop] (7.0, 5.25) -- (9.15, 5.25) -- (9.15, 1.45) -- (zb.east);
\draw[yloop] (7.0, 5.4) -- (9.3, 5.4) -- (9.3, 0.8) -- (7.0, 0.8) -- (yb.south);
\node[note, anchor=east, text=blue!55!black] (lL) at (9.05, 4.3) {$L\times$};
\node[note, anchor=east, text=orange!75!black] (l1) at (9.05, 3.9) {$1\times$};
\node[note, anchor=north] (hnote) at (7.0, 0.58) {$H$ cycles per step, gradient in the last};

\node[panel, fill=green!5, minimum width=2.8cm, minimum height=4.7cm, anchor=south] (dec) at (11.25, 1.45) {};
\node[ptitle] at (dec.north) {Decoder};
\node[dbox] (head) at (11.25, 5.05) {Output head\\[1.5pt] \scriptsize MIS: node scores\\[-1pt] \scriptsize TSP: edge scores};
\node[dbox, ours] (dcd) at (11.25, 3.625) {Decode \& select\\[1.5pt] \scriptsize MIS: greedy\\[-1pt] \scriptsize TSP: edge insertion};
\node[dbox, ours] (ver) at (11.25, 2.2) {Post-processing\\[1.5pt] \scriptsize MIS: peel $k_{\mathrm{p}}$, restart\\[-1pt] \scriptsize TSP: 2-opt};
\draw[arr] (head) -- (dcd);
\draw[arr] (dcd) -- (ver);
\draw[arr] (7.0, 5.55) -- (9.68, 5.55) node[note, pos=0.5, above=2pt] {$y$} -- (9.68, 5.05) -- (head.west);
\draw[arr] (ver.east) -- (12.75, 2.2) |- (13.0, 4.95);
\draw[arr] (ver.east) -- (12.75, 2.2) |- (13.0, 2.6);

\node[ptitle] at (13.6, 6.15) {Solution};
\misgraph{13.1}{4.45}{1}
\node[note] at (13.6, 4.05) {Independent\\ set};
\tspinst{13.1}{2.1}{1}
\node[note] at (13.6, 1.8) {Tour};

\draw[darr] (ver.south) -- (11.25, -0.2) -- (7.0, -0.2) -- (7.0, 0.2);
\node[onote, anchor=south] at (9.1, -0.17) {Train: halt once solved};
\node[onote, anchor=north] at (9.1, -0.23) {Self-relabel: better solution becomes label};
\path (enc.east); \pgfgetlastxy{\px}{\py} \typeout{PROBE fra enc.east=\px,\py}
\path (bb.west); \pgfgetlastxy{\px}{\py} \typeout{PROBE fra bb.west=\px,\py}
\path (bb.east); \pgfgetlastxy{\px}{\py} \typeout{PROBE fra bb.east=\px,\py}
\path (bb.north); \pgfgetlastxy{\px}{\py} \typeout{PROBE fra bb.north=\px,\py}
\path (dec.west); \pgfgetlastxy{\px}{\py} \typeout{PROBE fra dec.west=\px,\py}
\path (dec.east); \pgfgetlastxy{\px}{\py} \typeout{PROBE fra dec.east=\px,\py}
\path (stack.west); \pgfgetlastxy{\px}{\py} \typeout{PROBE fra stack.west=\px,\py}
\path (xin.west); \pgfgetlastxy{\px}{\py} \typeout{PROBE fra xin.west=\px,\py}
\path (sa.west); \pgfgetlastxy{\px}{\py} \typeout{PROBE fra sa.west=\px,\py}
\path (ver.east); \pgfgetlastxy{\px}{\py} \typeout{PROBE fra ver.east=\px,\py}
\path (head.west); \pgfgetlastxy{\px}{\py} \typeout{PROBE fra head.west=\px,\py}
\path (lL.east); \pgfgetlastxy{\px}{\py} \typeout{PROBE fra lL.east=\px,\py}
\path (l1.east); \pgfgetlastxy{\px}{\py} \typeout{PROBE fra l1.east=\px,\py}
\path (hnote.south); \pgfgetlastxy{\px}{\py} \typeout{PROBE fra hnote.south=\px,\py}
\path (Ebox.east); \pgfgetlastxy{\px}{\py} \typeout{PROBE fra Ebox.east=\px,\py}
\end{tikzpicture}

%% file: figures/recursion.tex
\begin{tikzpicture}[
  x=1cm, y=1cm, font=\footnotesize, >={Latex[length=1.3mm, width=1.1mm]}, line width=0.5pt,
  st/.style={draw=trmcol, rounded corners=2pt, minimum width=5.5mm, minimum height=5.5mm,
             fill=gray!12, inner sep=1pt},
  f/.style={draw=ourscol, line width=0.8pt, fill=gray!12, minimum width=5.5mm, minimum height=5.5mm, inner sep=1pt},
  grp/.style={draw=trmcol, rounded corners=4pt, inner sep=5pt},
  note/.style={font=\scriptsize, align=center, inner sep=1pt},
  gnote/.style={note, text=gray!30!black},
  onote/.style={note, text=ourscol},
  arr/.style={->, trmcol},
  oarr/.style={->, ourscol},
  darr/.style={->, dashed, trmcol},
]

\node[st] (e) at (0.5, 0.8) {$e$};
\node[gnote, below=1pt of e] {Tokens};
\node[f] (E) at (0.5, 2.2) {$A$};
\node[onote, above=1pt of E] {$+\gamma_h A$\\ in every $f_\theta$};

\node[st] (z0) at (1.9, 0.0) {$z$};
\node[f]  (f1) at (2.75, 0.0) {$f_\theta$};
\node[st] (z1) at (3.6, 0.0) {$z$};
\node[f]  (fL) at (5.1, 0.0) {$f_\theta$};
\node[gnote, inner sep=0pt, yshift=-0.24pt] (dots1) at ($(z1.center)!0.5!(fL.center)$) {$\cdots$};
\node[st] (zL) at (5.95, 0.0) {$z$};
\node[st] (y0) at (2.75, 1.6) {$y$};
\node[f]  (fy) at (5.95, 1.6) {$f_\theta$};
\node[st] (y1) at (6.8, 1.6) {$y$};
\draw[arr] (z0) -- (f1); \draw[arr] (f1) -- (z1); \draw[trmcol] (z1) -- (dots1);
\draw[arr] (dots1) -- (fL); \draw[arr] (fL) -- (zL);
\draw[arr] (y0) -- (f1);
\draw[arr] (y0.east) -| (fL.north);
\draw[trmcol] (e.east) -- (5.83, 0.8) arc (180:0:0.12) -- (9.25, 0.8);
\fill[trmcol] (2.75, 0.8) circle (0.9pt); \fill[trmcol] (5.1, 0.8) circle (0.9pt);
\draw[arr] (zL) -- (fy);
\draw[arr] (y0.east) -- (fy.west);
\draw[arr] (fy) -- (y1);
\node[gnote, anchor=north] at (4.6, -0.6) {$L$ latent updates};
\begin{scope}[on background layer]
  \node[grp, dashed, fit=(z0)(f1)(z1)(dots1)(fL)(zL)(y0)(fy)(y1),
        label={[gnote]above:Cycles $1$ to $H{-}1$, no gradient}] (c1) {};
\end{scope}

\node[st] (zH) at (8.4, 0.0) {$z$};
\coordinate (padH) at (8.0, 0.0);  
\node[f]  (fH) at (9.25, 0.0) {$f_\theta$};
\node[st] (yH0) at (9.25, 1.6) {$y$};
\node[f]  (fyH) at (10.5, 1.6) {$f_\theta$};
\node[gnote, inner sep=0pt, yshift=-0.24pt] (dotsH) at ($(fH.east)!0.5!(fyH.center |- fH.east)$) {$\cdots$};
\node[st] (yH) at (11.35, 1.6) {$y$};
\draw[arr] (zH) -- (fH); \draw[trmcol] (fH) -- (dotsH); \draw[arr] (dotsH) -| (fyH.south);
\draw[arr] (yH0) -- (fH);
\draw[arr] (yH0) -- (fyH);
\draw[arr] (fyH) -- (yH);
\begin{scope}[on background layer]
  \node[grp, fit=(padH)(zH)(fH)(dotsH)(yH0)(fyH)(yH),
        label={[gnote]above:Cycle $H$, with gradient}] (cH) {};
\end{scope}
\fill[trmcol] (9.25, 0.8) circle (0.9pt);
\node[gnote, inner sep=1pt, yshift=-0.24pt] (dotsY) at ($(c1.east |- 0,1.6)!0.5!(cH.west |- 0,1.6)$) {$\cdots$};
\node[gnote, inner sep=1pt, yshift=-0.24pt] (dotsZ) at ($(c1.east |- 0,0)!0.5!(cH.west |- 0,0)$) {$\cdots$};
\draw[trmcol] (y1) -- (c1.east |- y1); \draw[arr] (cH.west |- yH0) -- (yH0);
\draw[trmcol] (zL) -- (c1.east |- zL); \draw[arr] (cH.west |- zH) -- (zH);

\node[st] (g) at (12.3, 1.6) {$g_\theta$};
\draw[arr] (yH) -- (g);
\node[draw=trmcol, rounded corners=2pt, fill=gray!12, align=center, inner sep=3pt, font=\scriptsize] (loss) at (13.85, 2.4) {Train: CE loss\\ vs.\ label $\ell$};
\node[draw=ourscol, line width=0.8pt, rounded corners=2pt, fill=gray!12, align=center, inner sep=3pt, font=\scriptsize] (dec) at (13.85, 0.8) {Test: decode $s$,\\ score with $c(s)$};
\draw[arr] (g.east) -- (12.75, 1.6) |- (loss.west);
\draw[oarr] (g.east) -- (12.75, 1.6) |- (dec.west);
\node[onote, below=1pt of loss] (halt) {Halt once solved};
\node[onote, anchor=north] (run) at (13.85, 0.42) {Run $D$ steps,\\ keep best $s$};

\node[gnote, anchor=north west, align=left] at (1.6, -0.6) {Test: $z + \sigma\xi$\\ $K$ rollouts};

\draw[darr] (cH.south) -- (cH.south |- 0,-1.45) -- (1.1, -1.45) -- (1.1, 0.0) -- (c1.west |- z0.west);
\node[gnote, anchor=north] at ($(1.1, -1.55)!0.5!(cH.south |- 0,-1.55)$) {Carry $(y, z)$ detached to step $t{+}1$};
\end{tikzpicture}